\documentclass{optica-article}

\journal{opticajournal} 

\articletype{Research Article}

\usepackage{xcolor}
\usepackage{subcaption}
\newcommand*{\Fig}[1]{{Fig.\ref{fig:#1}}}
\newcommand*{\Table}[1]{{Table.\ref{table:#1}}}
\newcommand{\SPACE}{\par\vspace{2mm}}

\begin{document}

\title{Cost-efficient Active Illumination Camera For Hyper-spectral Reconstruction}

\author{Yuxuan Zhang\authormark{1}, T.M. Sazzad, Yangyang Song,
Spencer J. Chang, Ritesh Chowdhry, Tomas Mejia, Anna Hampton, Shelby Kucharski,
Stefan Gerber, Barry Tillman, Marcio F. R. Resende, William M. Hammond, Chris H. Wilson, Alina Zare
and Sanjeev J. Koppal\authormark{2}}

\address{
\authormark{1, 2} Yuxuan Zhang and Sanjeev J. Koppal are with the Department of Electrical and Computer Engineering, University of Florida. Sanjeev J. Koppal holds concurrent appointments as an Associate Professor of ECE at the University of Florida and as an Amazon Scholar. This paper describes work performed at the University of Florida and is not associated with Amazon.
}

\email{\authormark{1}zhangyuxuan@ufl.edu}
\email{\authormark{2}sjkoppal@ece.ufl.edu}


\begin{abstract*}
Hyper-spectral imaging has recently gained increasing attention for use in different applications, including agricultural investigation, ground tracking, remote sensing and many other. However, the high cost, large physical size and complicated operation process stop hyperspectral cameras from being employed for various applications and research fields. In this paper, we introduce a cost-efficient, compact and easy to use active illumination camera that may benefit many applications.
We developed a fully functional prototype of such camera. With the hope of helping with agricultural research, we tested our camera for plant root imaging. In addition, a U-Net model for spectral reconstruction was trained by using a reference hyperspectral camera's data as ground truth and our camera's data as input. We demonstrated our camera's ability to obtain additional information over a typical RGB camera. In addition, the ability to reconstruct hyperspectral data from multi-spectral input makes our device compatible to models and algorithms developed for hyperspectral applications with no modifications required.
\end{abstract*}

\section{Introduction}

Hyper-spectral imaging has gained increasing interest in recent years. The ability to capture rich spectral information beyond traditional RGB channels has been proven to help reveal more information from the subject of interest, including agricultural and plant researches \cite{amigo2015hyperspectral, MISHRA201749, behmann2014detection, amigo2013hyperspectral, chang2023hyperpri}. Compared to RGB cameras, hyper-spectral imaging provides both richer information inside the visible bandwidth (380 - 700nm) and additional information outside of the visible bandwidths which would be unavailable otherwise.
However, hyper-spectral cameras also tend to cost significantly more than traditional cameras and to have a large physical size. Its shortcomings limited the chance of integrating hyper-spectral imaging into a larger variety of industrial applications or research projects.

As an example, minirhizotron is a commonly used tool for research of plant roots \cite{boris2013, sharma2024, postic2019, johnson2001}. It is a tube shaped inspection tool made of transparent materials. Its inner diameter varies from as small as $5$cm up to $20-30$cm. It reaches deep into soil near the plant to be inspected, enabling researchers to visually inspect plant roots in a nondestructive way. A lot of studies have been carried out with the help of minirhizotron and compact monochrome or RGB cameras \cite{rajurkar2022, liedgens2001, chen2018}. Although these studies could potentially be significantly benefited from hyper-spectral imaging, up to our best knowledge, there does not exist a feasible way to provide hyperspectral imaging inside such narrow space.

Given the potential benefits of enabling hyper-spectral imaging inside narrow spaces such as a minirhizotron, we investigated a new approach of obtaining spectral information. Instead of performing a full-spectral scan, our approach focuses on a few discrete bands that were selected to carry the most potential information. Our device utilizes readily available parts from the consumer market, and our design features an ultra compact size so it can fit into applications with very narrow operating space for imaging. In addition to compact size, our device also has the advantages of low cost and ease of use. Such advantages makes it possible to be massively deployed and autonomously operated.

\section{Related Work}

Hyper-spectral imaging (HSI) have had a significant impact in agriculture and remote sensing. For example, classification allows for separating diseased from healthy plants using additional color band information~\cite{ahmad_fast_2022, hu_lightweight_2021, chen_deep_2014, chen_hyperspectral_2011, el_rahman_hyperspectral_2016}. In addition, detecting the maturity of crops and finding stressed crops are important HSI tasks \cite{gao_real-time_2020, varga_measuring_2021, zou_peanut_2019, nguyen_early_2021, aredo_predicting_2019, behmann_detection_2014}. 

The ever-increasing accessibility of HSI has fueled widespread impact\cite{Khan,Li,Yuan,Uzair,Tang,Lu}, with studies on hundreds of spectral bands \cite{sellami2018hyperspectral,Sun,zheng2022effective,yuan2016discovering}. These bands provide rich information that can be analyzed and identified to differentiate objects in the scene \cite{Wang,wei2017applications}. HSI have impacts on fields as diverse as agriculture (plant disease detection and classification, etc), tissue analysis in medical science, forest land detection, mining, and mineral study, vegetation estimation, protection of the environment, biological analysis, etc \cite{He,Van,Luo,akbari2010detection,liu2019review,zheng2021generalized}.

However, the large data footprint of the spectral bands has disadvantages, such as high computational time complexity, transmission, storage and analysis \cite{zheng2022effective,xie2019unsupervised,zhang2020deep,yang2011unsupervised,zheng2022rotation}. Therefore, redundant information and time complexity minimization is important to reconstruct hyper-spectral data efficiently \cite{zhang2020deep,wang2019hyperspectral,sawant2020hyperspectral,yang2010efficient}. Techniques for compressing the measurements include clustering \cite{sawant2020unsupervised,jia2012unsupervised}, Ranking approaches \cite{ghorbanian2020clustering,martinez2007clustering,song2019class,chang2006constrained}, greedy approaches \cite{haut2019visual,mao2004orthogonal} and evolutionary approaches \cite{yuan2014hyperspectral,tu2021feature}. 

\textbf{Dimensionality reduction of HS samples:} A range of band-selection approaches have been proposed in the existing literature. Typically, to decrease the dimensionality of a hyper-spectral image, these band selection approaches can be categorised into two specific groups that include feature extraction \cite{lupu2022stochastic,wang2020dimensionality} and feature selection (band selection) \cite{zhai2018laplacian,jia2015novel,feng2016multiple,yuan2015dual,wang2019hyperspectral,zhang2018geometry}. Interestingly, the above mentioned two specific methods extract or select data from all HSI bands to correspond to the whole spectral cube, and the outcomes are almost or roughly equivalent to the full HSI bands. Currently, traditional feature selection approaches are considered to be the most commonly applied techniques which include Principal Component Analysis (PCA) \cite{song2019class}, Maximum Noise Fraction (MNF) \cite{sun2022hyperspectral}, Genetic Algorithm, and FICA (Fast Independent Component Analysis) \cite{lupu2022stochastic}. For feature extraction approaches, high dimensional spatial data are mapped into low-dimensional space considering specific criteria, thus extracting a complete new sub-set of features which represents the original HS data. Unfortunately, during spatial transformation, the physical denotation of the original HS data cannot be found same as well as it is also possible that some of the key or main information may be lost forever. On the other hand, for the band selection approaches a distinct and representative sub-set is selected from the unique hyper-spectral data which preserves the physical denotation and information without loss. Additionally, it also preserves the intrinsic characteristics of the HS data. In this article, we have focused on feature (band) selection rather than feature (band) extraction.

\textbf{Supervised and Semi-supervised labelling:} Supervised approaches require labelled samples for the selection of most favourable bands during training and learning \cite{wang2018locality} where similarity measure metric is used among the class labels. These techniques require a number of assessment conditions which can be categorised as: information divergence \cite{sun2022novel}, maximum ellipsoid volume \cite{geng2014fast}, Euclidean distance \cite{wang2016salient} etc. For semi-supervised approaches graph based models are used for labelled and unlabelled data samples for the selection of appropriate spectral bands \cite{gao2017additional,sun2014exemplar} but suffers from providing contextual information.

There exist attempts to integrate multispectral camera into minirhizotron\cite{rahman2020soilcam}. This work implements an automatically operated multi-spectral camera that can actuate by itself inside a minirhizotron. Their camera was also equipped with multi-bandwidth light sources. However, instead of exploring the information contained in data acquired from such device, this work primarily focused on autonomous operation and remote deployment.




\section{Hardware Design}


We propose a type of camera that uses LEDs of different bands instead of color filters to capture spectral information. In our setup, the camera sensor does not need to be equipped with any type of optical filter, not even the Bayer filter for an RGB sensor or IR filter commonly seen on the back of lenses. This setup works perfectly in the scenarios where the on-camera LEDs are the only source of light. Moreover, it can also deal with static scenes with a moderate amount of ambient light by treating the ambient light as "dark field" and only tracking the incremental light upon firing each LED.

\begin{figure}[!t]
\centering
\subfloat[]{
\includegraphics[width=0.32\textwidth]{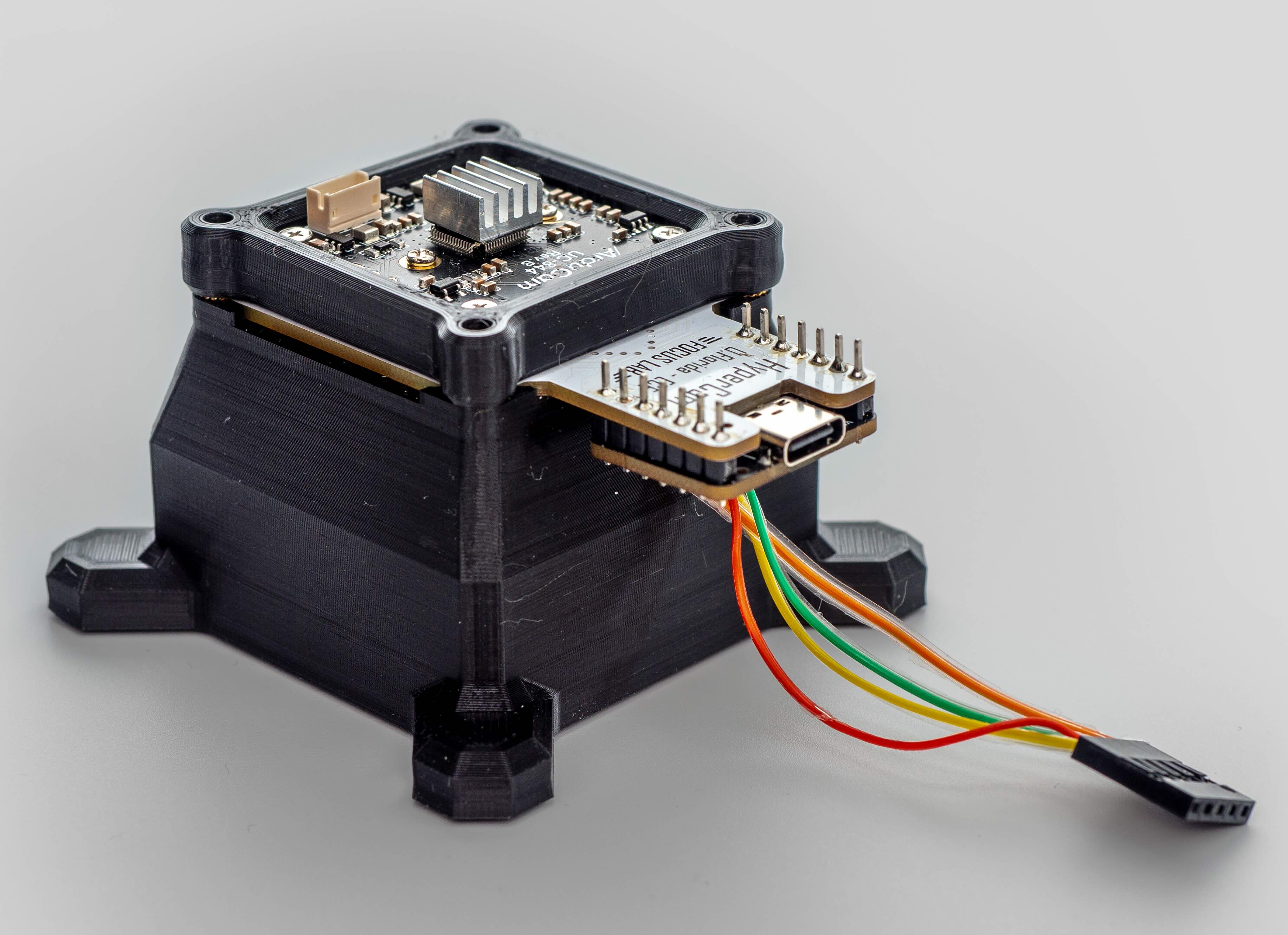}
\label{fig:v0-camera}
}
\subfloat[]{
\includegraphics[width=0.67\textwidth]{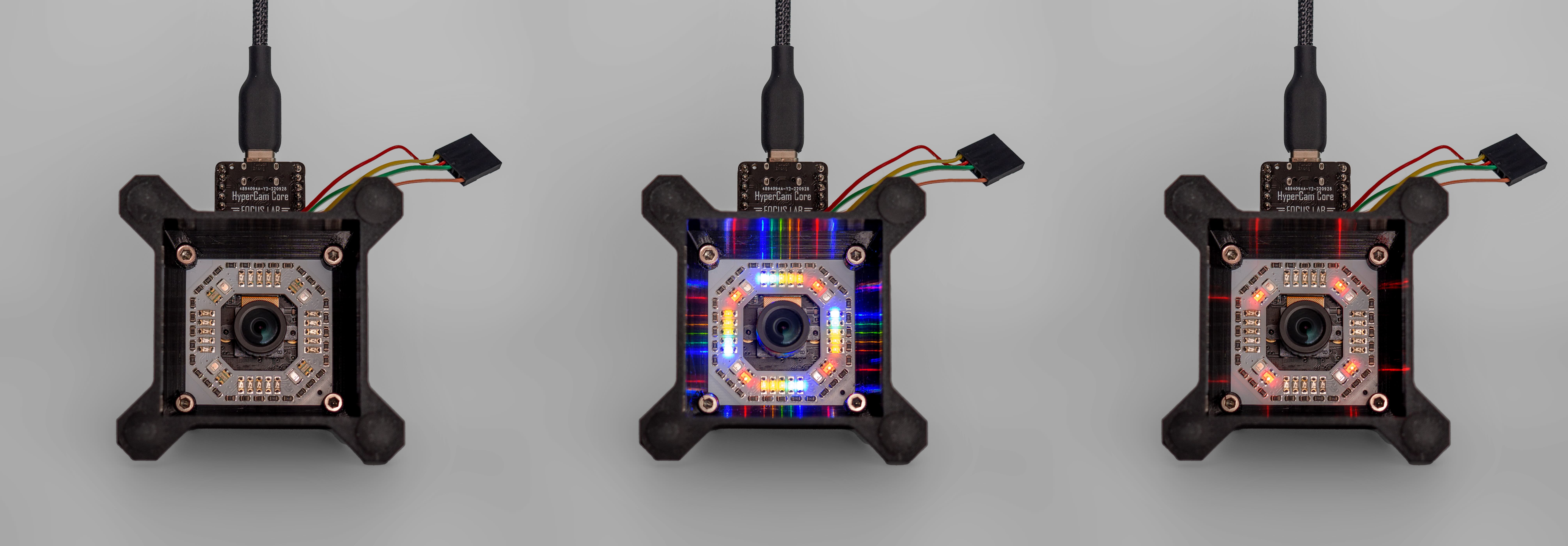}
\label{fig:led-layout}
}
\caption{
(a) Photo of Version-0 camera designed for data collection;
(b) Layout of the LEDs that minimizes the intensity distribution of
different colors.
Left: The LED module;
Center: All LEDs turned on, excluding UV;
Right: Single color lit.}
\end{figure}

Our prototype was designed to house 8 different types of LEDs, each having a unique bandwidth of our choice. The optical characteristics of each LED model is listed in \Table{LED_Spec}, the selection of bandwidths will be further discussed in next section. We mount 4 LEDs per color, and as is shown in \Fig{led-layout}, the LEDs with same color are mounted in a rectangular pattern in order to illuminate the subject as evenly as possible. Such arrangement minimizes the difference of light distribution across different colors.

Since all parts of our camera are standard market-ready parts, we can achieve an extremely low unit cost compared to hyper-spectral cameras currently in the market. In addition, the constraints of the environment in which the camera will be operated is another crucial aspect we should take care of. In our case, we are targeting the minirhizotrons broadly used to study plants. Minirhizotron is a specialized tool used in biology and agriculture researches to study roots in their natural environment. It is a transparent tube inserted into the soil, making it possible to observe roots as they grow over time.

The challenge lies in the geometry of the minirhizotron. The diameter of a typical minirhizotron is between 5 to 10 centimeters. And the one used by our collaborating research group is only 5cm in diameter. Even for our camera, fitting itself into such a narrow tube is a significant engineering challenge. Most of the existing hyper-spectral cameras in the market feature a much larger dimension, making them impossible to be deployed in such environments.

\begin{figure}
\centering
\includegraphics[width=0.6\columnwidth]{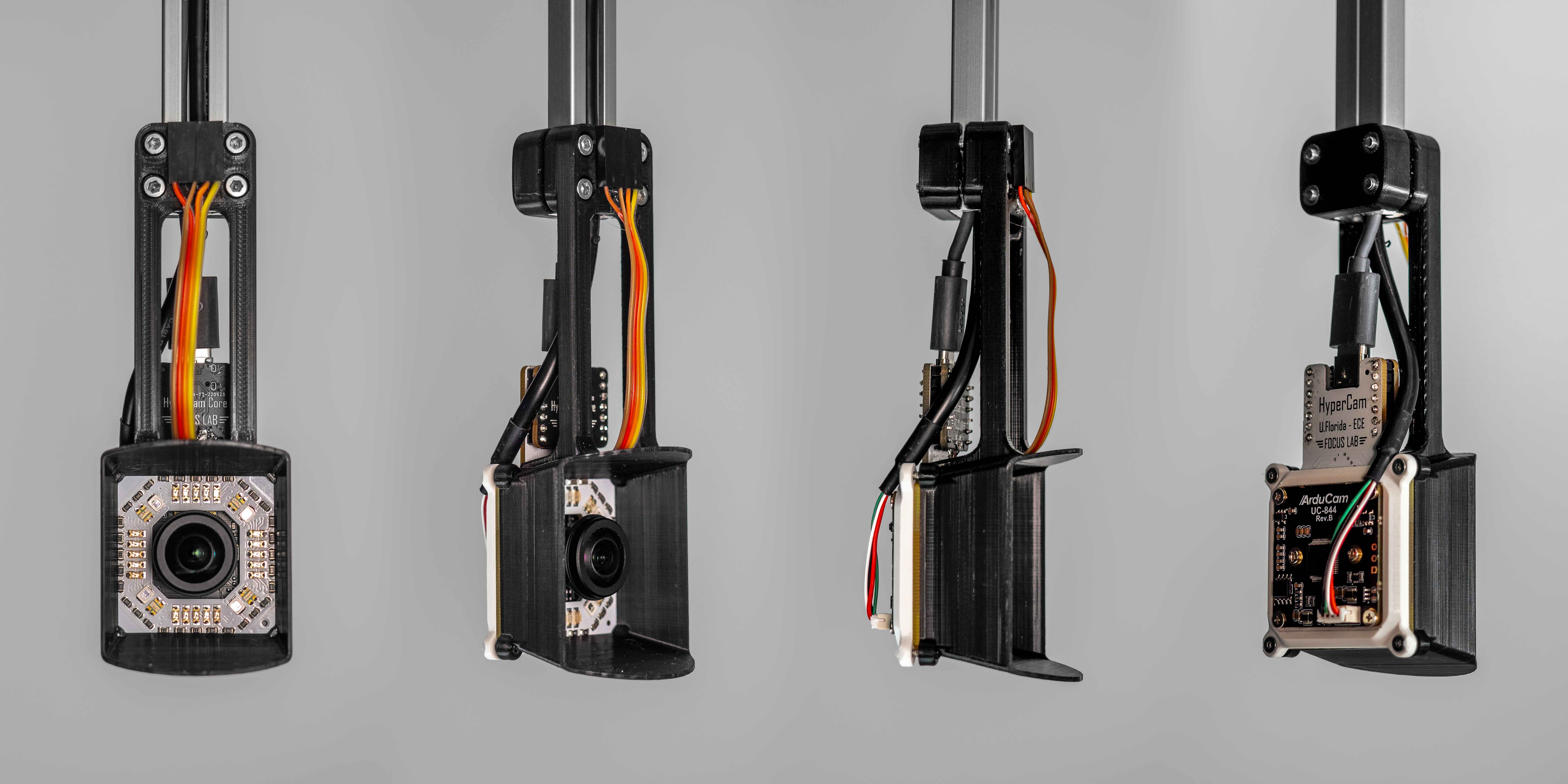}
\SPACE
\caption{The same components can fit into a minirhizotron with a dedicated case design.}
\label{fig:camera-v2}
\end{figure}

Our design addressed these challenges by both utilizing small monochrome sensor modules and a custom designed printed circuit board (PCB) to integrate the LEDs. The camera module and the LED module were stacked upon each other with only 5mm distance. The LED module has an opening in its center to let through a wide angle lens that sits on the camera module's PCB. Each module has an upstream USB port to be connected with the host computer (a raspberry pi), and the USB port handles both power and communication. The first camera was designed to work with the flat surface of a rhizobox, but the same core components can be easily adapted into a 5cm diameter minirhizotron with a different case (\Fig{camera-v2}).

\section{Selection of Bandwidths}

\begin{table}[!t]
\centering
\caption{Characteristics of Selected LEDs}
\label{table:LED_Spec}
\SPACE
\begin{tabular}{c|cccc}
\hline
Type & $\lambda_\text{peak}$ & $\Delta\lambda$ & $V_\text{F}$ & $I_\text{max}$\\
\hline
Ultra Violet & $395$ nm & $10$ nm & $3.3$ V & $~60$ mA\\
Blue     & $466$ nm & $15$ nm & $2.9$ V & $~30$ mA\\
Green    & $520$ nm & $15$ nm & $2.9$ V & $~30$ mA\\
Yellow-Green  & $573$ nm & $20$ nm & $2.4$ V & $~25$ mA\\
Yellow   & $585$ nm & $20$ nm & $2.4$ V & $~25$ mA\\
Orange   & $600$ nm & $20$ nm & $2.4$ V & $~25$ mA\\
Red      & $660$ nm & $17$ nm & $2.1$ V & $100$ mA\\
Infrared & $940$ nm & $40$ nm & $1.3$ V & $200$ mA\\
\hline
\multicolumn{5}{l}{$\lambda_\text{peak}$ is the peak wavelength of the spectrum;} \\
\multicolumn{5}{l}{$\Delta\lambda$ is the half width of the spectrum.}
\end{tabular}
\end{table}

We performed analysis on data gathered by the reference HSI camera, using the same rhizobox samples
that will be used by our camera (analysis was performed by T.\thinspace M.\thinspace Sazzad). Out of all 8 bandwidths,
three of our selections matched exactly with the optimal bands (blue, green and red). The other bandwidths
are limited by stock availability when producing the prototype, and thus did not match exactly with calculated
optimal values.

\section{Data Collection and Processing}


We collected a dataset to train the U-Net based model for spectrum reconstruction.
Since the ground truth was required to train the model, and our reference data needed to be captured by a reference HSI device which does not fit into a minirhizotron, the initial version of camera we developed (shown in \Fig{v0-camera}) was especially designed to match the results from the reference camera. Both cameras would take hyper-spectral images of plant roots in rhizoboxes. The rhizobox is a container that works similarly to a minirhizotron, the difference being the rhizobox features a flat transparent surface so it's easier to take photos with all types of cameras. In addition, the compact size of a rhizobox makes it easy to be moved around.

\begin{figure}
\centering
\subfloat[][Original]{
\includegraphics[width=0.33\textwidth]{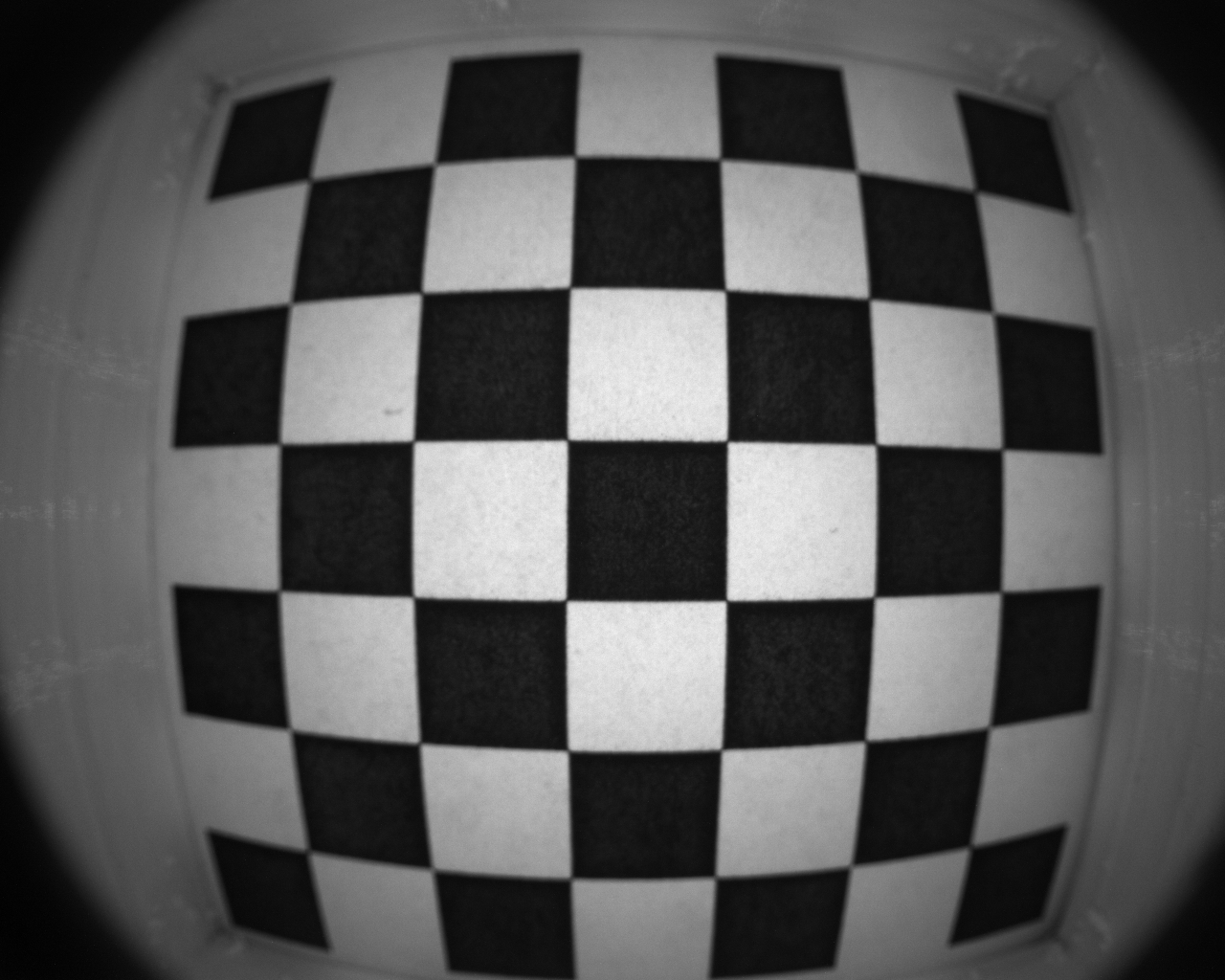}
}
\subfloat[][Intensity Normalized]{
\includegraphics[width=0.33\textwidth]{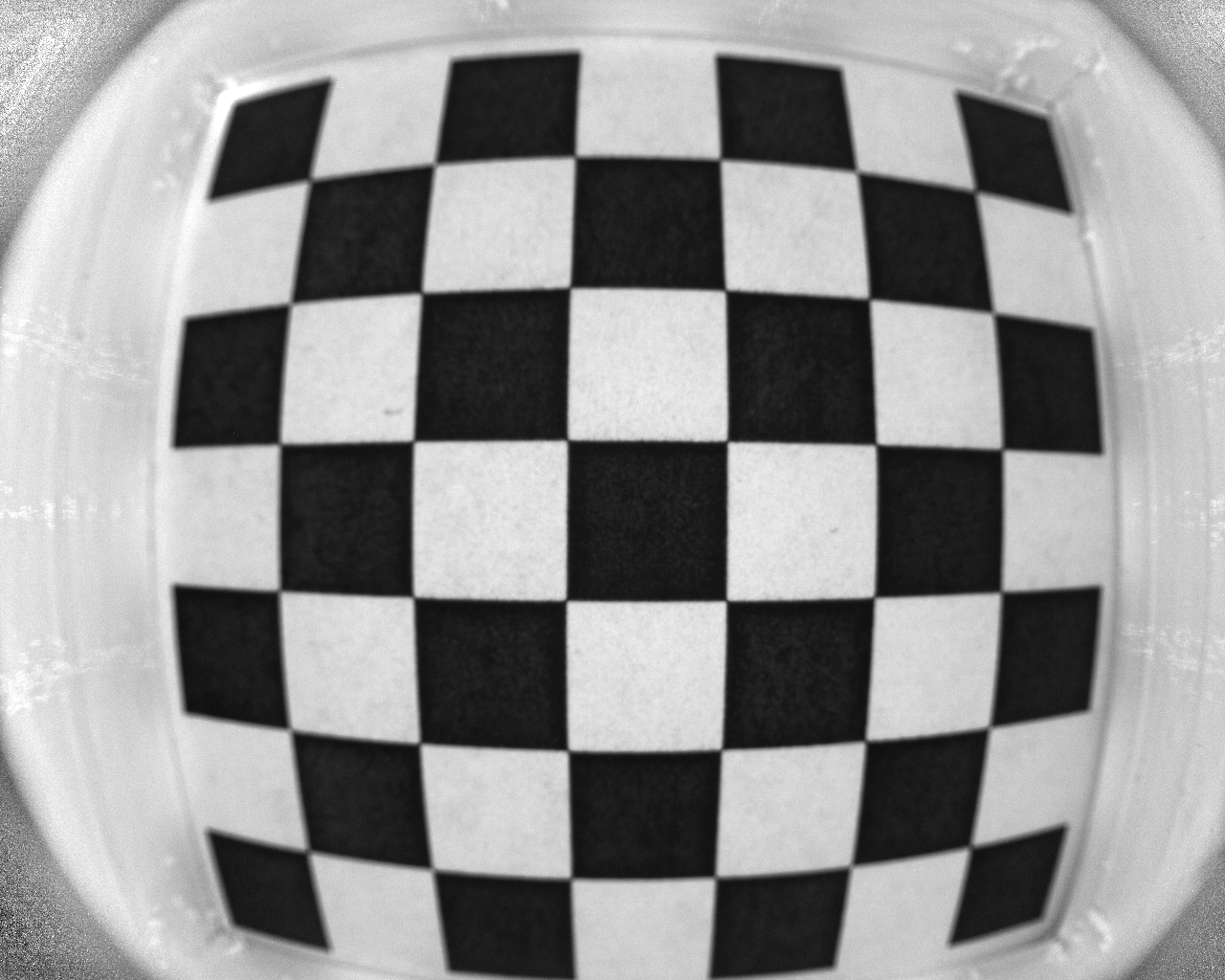}
}
\subfloat[][Undistorted]{
\includegraphics[width=0.264\textwidth]{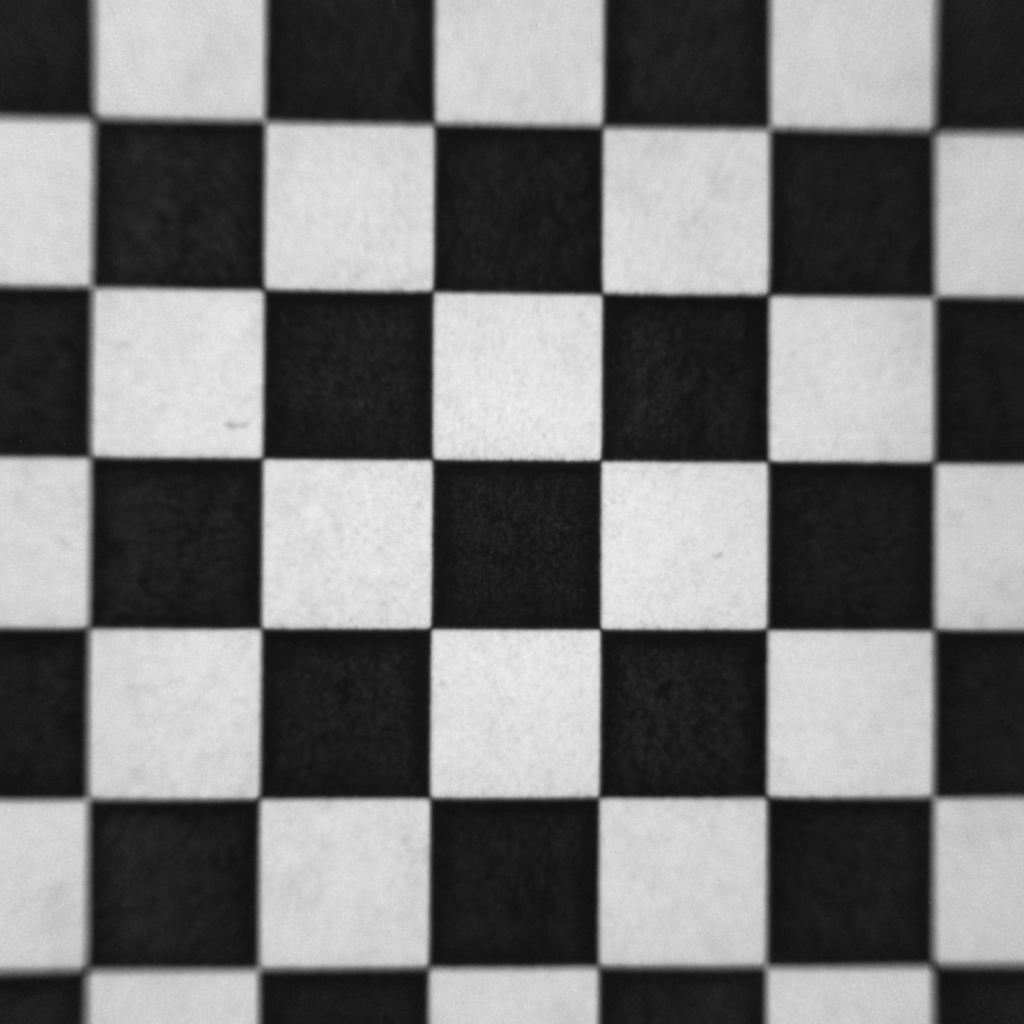}
}

\caption{Calibration pipeline for our custom camera.}
\label{fig:calibration-pipeline}
\end{figure}

\begin{figure}
\centering
\subfloat[][Raw image captured by our camera]{
\includegraphics[width=\columnwidth]{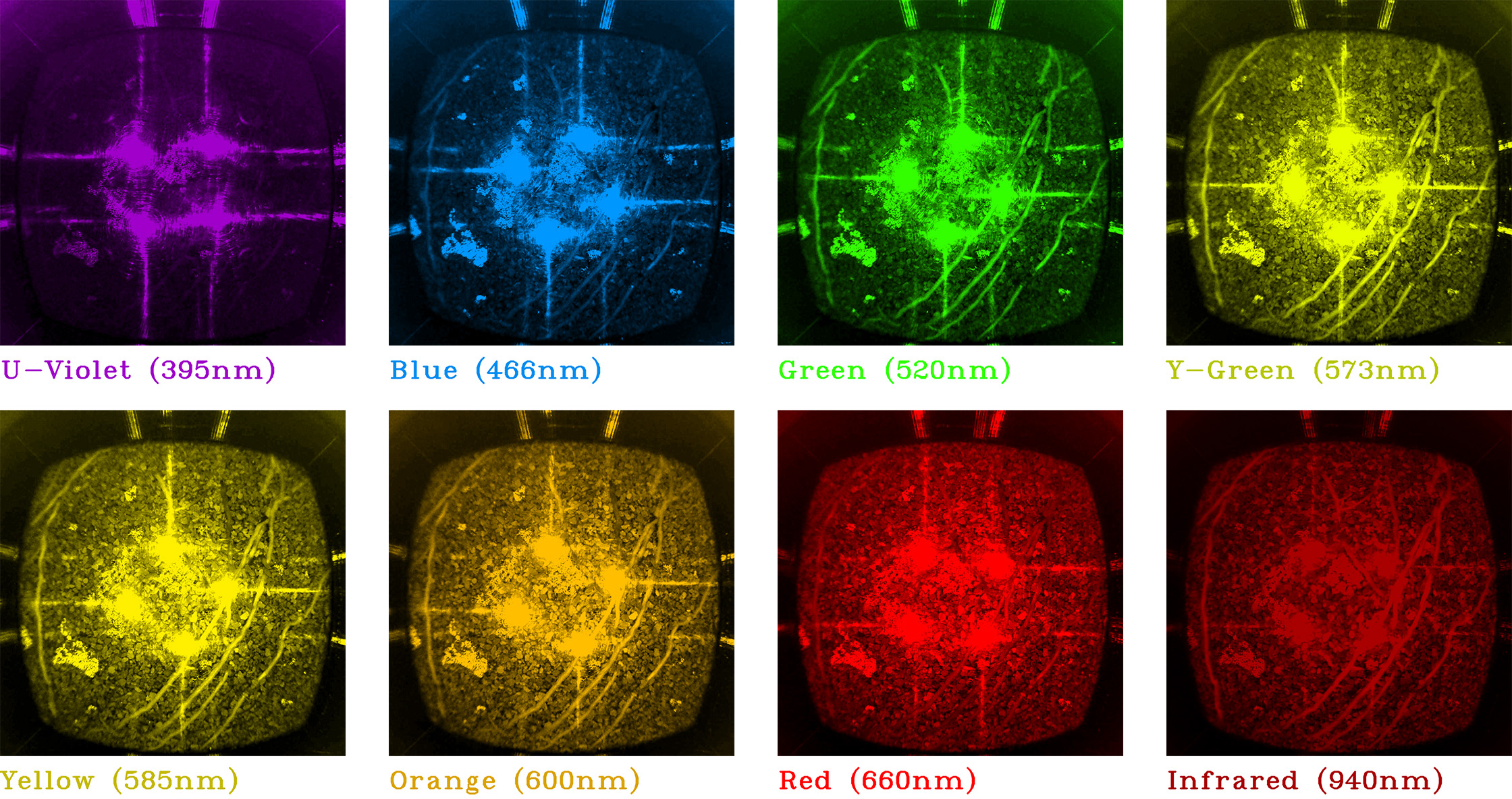}
\vspace{2mm}
\label{fig:raw-image-our-camera}
} \\ \vspace{10mm}
\subfloat[][Results of post-processing pipeline]{
\includegraphics[width=\columnwidth]{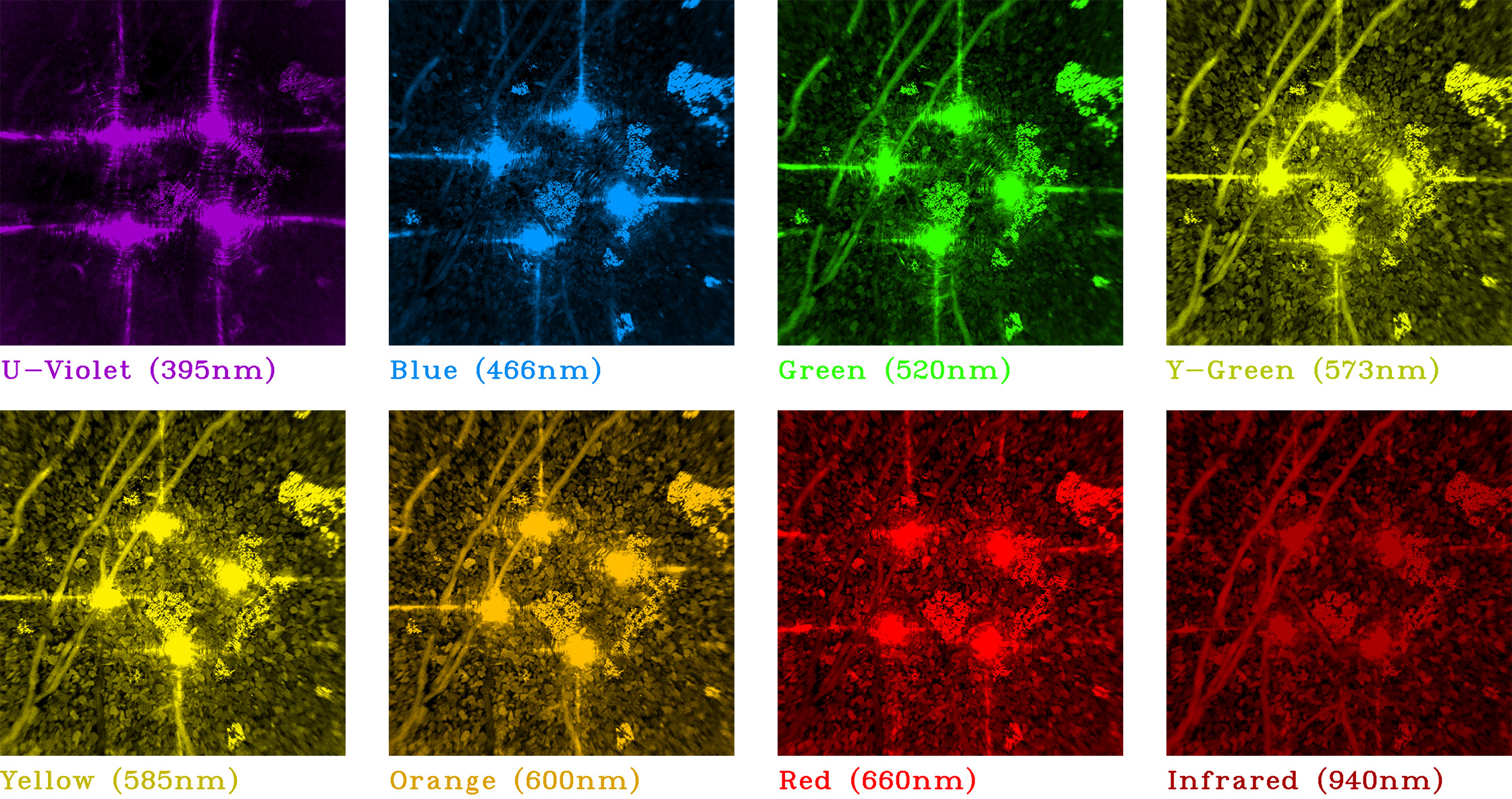}
\vspace{2mm}
\label{fig:cal-image-our-camera}
} \\ \vspace{2mm}
\caption{
(a) Raw image captured by our camera, showing 8 different bands of a same region, each rendered with corresponding pseudo color;
(b) Processed post-processing results, all calibrations applied. The distortions are corrected, camera frames are clipped out of view, and the dark corners are adjusted so that the brightness is uniform across the entire image.
}
\end{figure}

Due to the distortion and uneven intensity distribution introduced by hardware,
the raw data (\Fig{raw-image-our-camera}) captured from our camera has to be
pre-processed before proceeding.
In our calibration pipeline, the raw image is firstly mapped by a "reference white"
calibration image, eliminating uneven intensity distribution introduced by
light sources and lens projection. The reference white was captured and applied
separately for each color.
Geometric correction was then applied based on a chess board image, which is
captured by the same camera. The brightness distribution calibrated in the previous
step will be retained during this process (calibration pipeline shown in
\Fig{calibration-pipeline}, sample calibrated image shown in
\Fig{cal-image-our-camera}).

\begin{figure}
\centering
\includegraphics[width=0.6\columnwidth]{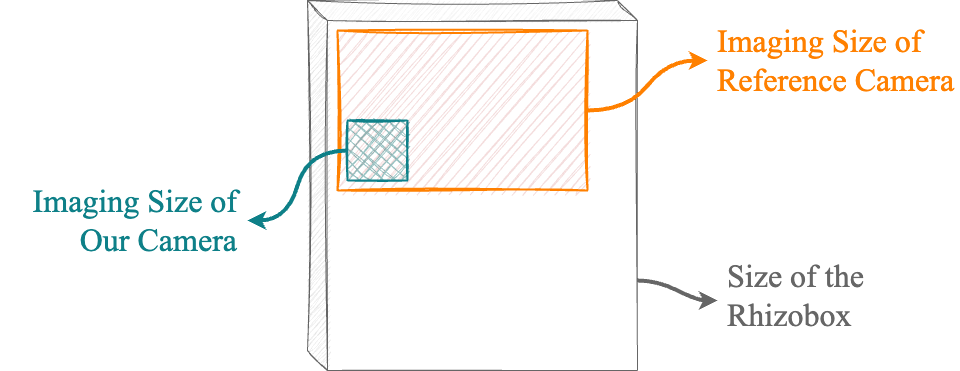}
\SPACE
\caption{Comparison of the sizes of the rhizobox, the reference camera's
imaging area, and our camera's imaging area. Dimensions shown in the figure are
conceptual, and are not strictly proportional to their actual sizes.}
\label{fig:frame-comparison}
\end{figure}

\begin{figure}
\centering
\subfloat[][Raw HSI image (Ground Truth)]{
\label{fig:align-A}\includegraphics[width=0.9\columnwidth]{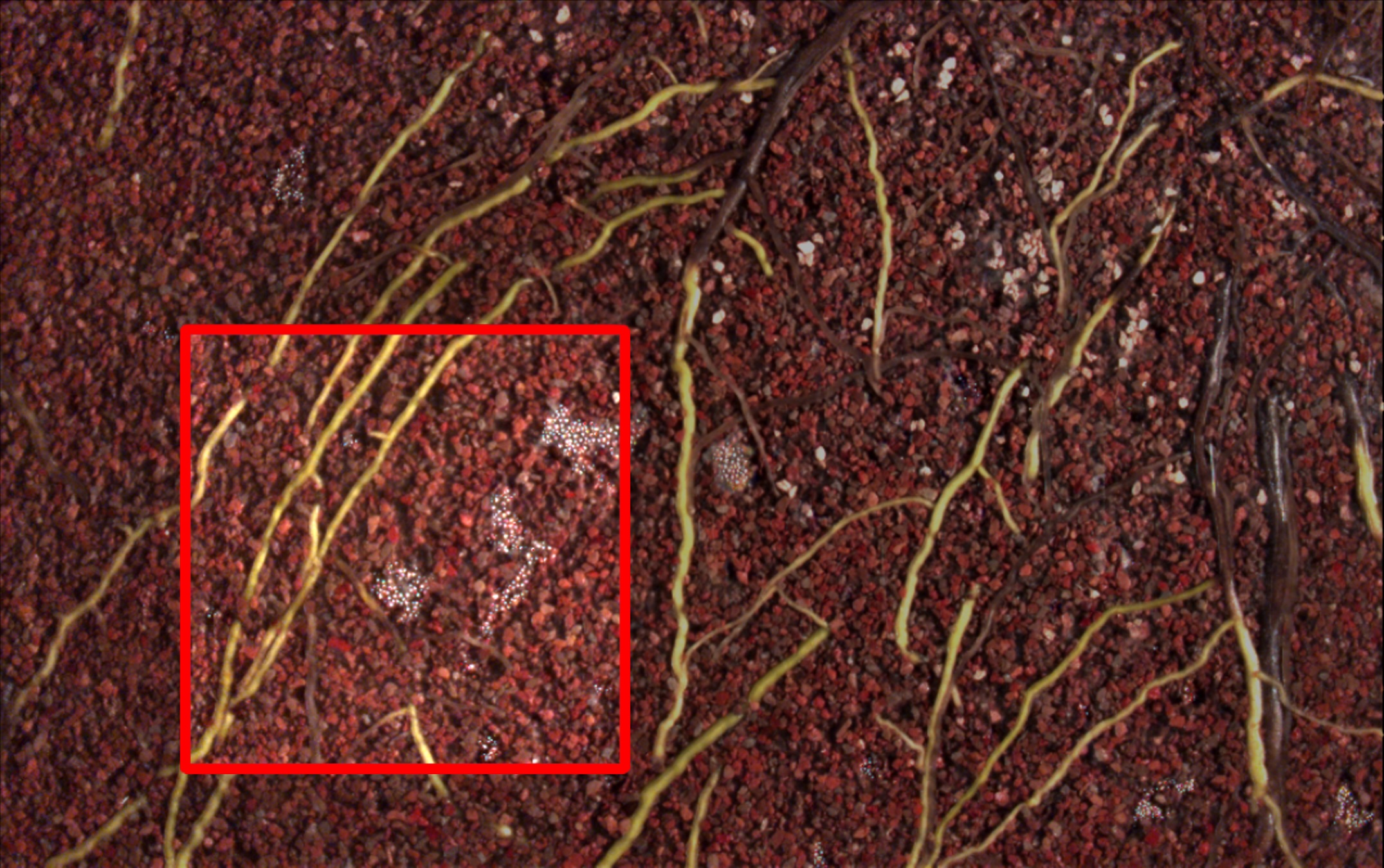}
}
\SPACE
\subfloat[][Matched Reference Image]{
\label{fig:align-B}\includegraphics[width=0.275\columnwidth]{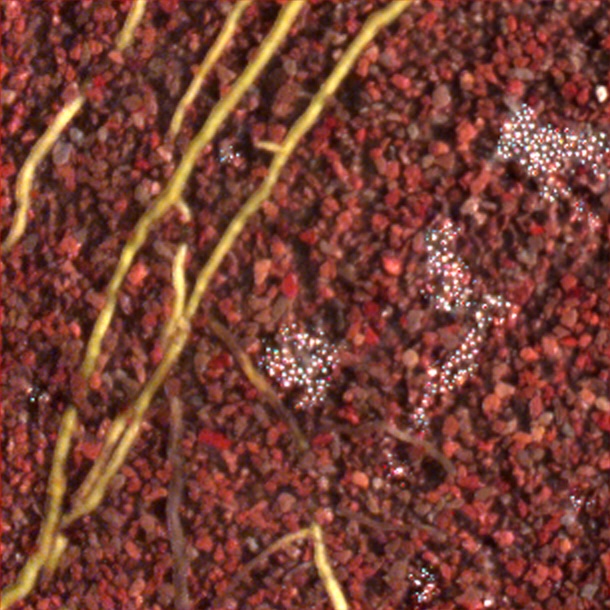}
}
\hspace{0.02\columnwidth}
\subfloat[][Our Camera's Result]{
\label{fig:align-C}\includegraphics[width=0.275\columnwidth]{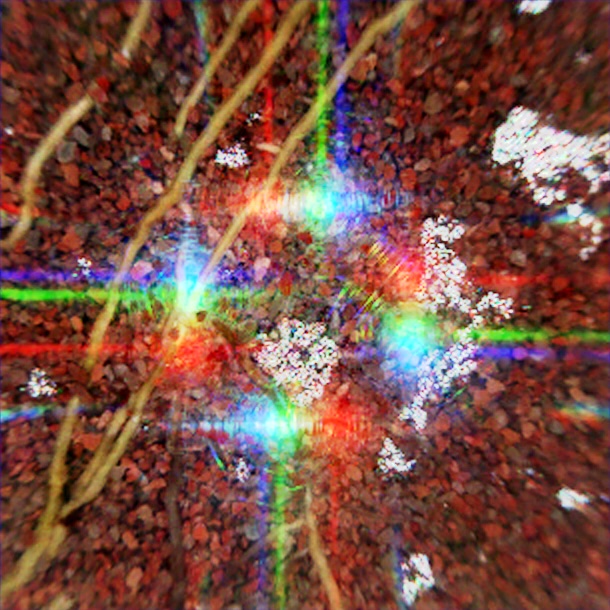}
}
\hspace{0.02\columnwidth}
\subfloat[][Align Mismatch]{
\label{fig:align-D}\includegraphics[width=0.275\columnwidth]{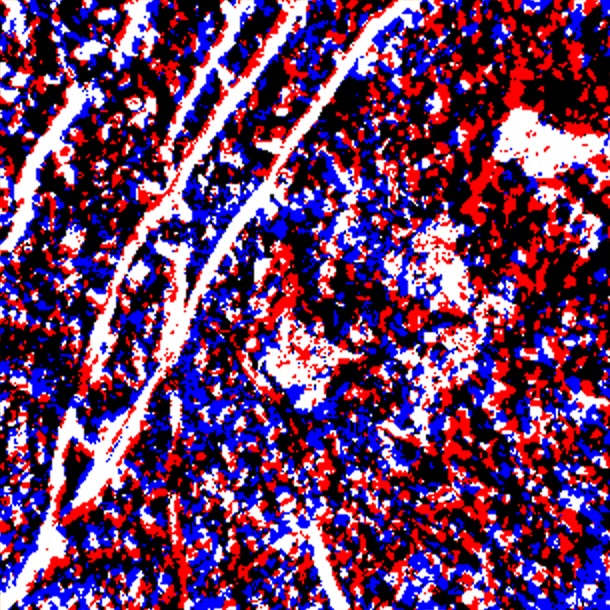}
}
\SPACE
\caption{Sample of a successful match between reference camera and our camera.
}
\label{fig:aligned-image}
\end{figure}

The final step of data processing is matching our camera's result with the
reference camera's result. The sizes of (1) the rhizobox, (2) the reference
camera's imaging area, and (3) our camera's imaging area are shown in
\Fig{frame-comparison}. We first manually downscale the size of our camera's
output ($1024 \times 1024$ after calibration) to match the pixel density of the
reference camera ($286 \times 286$). Then we are able to run template matching
on the reference image (using the down-scaled image as template).
\Fig{aligned-image} shows a sample of a successful match.

\section{Reconstruction Model}

A model based on U-Net was implemented and trained using the data we
collected. The model was intended for both removing the bright LED light spots introduced by our custom light module and expanding the number of bands from 8 to 299.

\subsection{Model Architecture}

U-Net is a convolutional neural network architecture that is commonly used for segmentation tasks in computer vision \cite{ronneberger2015}. However, instead of performing typical segmentation tasks, we explored the feasibility of using such architecture to perform \textbf{spectral reconstruction} tasks. This is based on the assumption that the U-Net could learn the inherent spectrum structure of different soil and root components based on the training set, and then transfer its prior knowledge to new data while maintaining contextual awareness.

\begin{figure}[ht]
\centering
\includegraphics[width=\columnwidth]{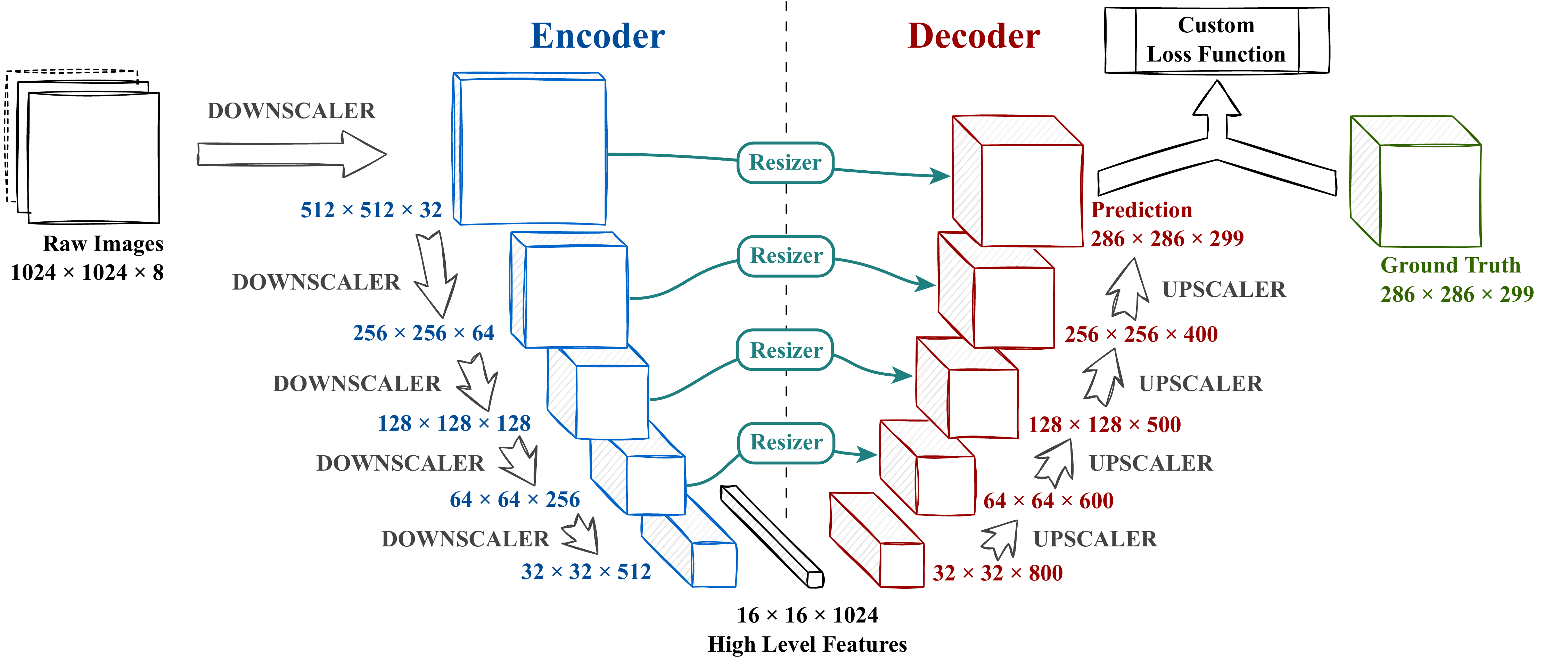}
\SPACE
\caption{Asymmetric network structure designed for spectral reconstruction.}
\label{fig:training-process}
\end{figure}

In order to fully leverage the resolution of our camera sensor, we reformed the model into an asymmetric structure. The left side of the network (encoder) descends faster on spatial resolution, while the right side (decoder) descends slower on spectral dimension. In this setup, the dimensions of the corresponding layers on each side no longer match. A resizer had to be implemented within each feed-forward shortcut connection.

\subsection{Data Augmentation}

\begin{figure}[ht]
\includegraphics[width=\textwidth]{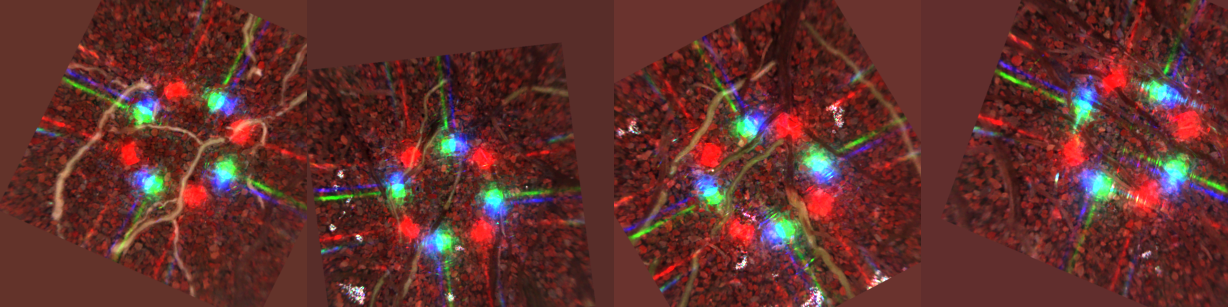}
\SPACE
\caption{Samples of random affine augmentation. Data points are shown in pseudo RGB.}
\label{fig:affine}
\end{figure}

\begin{table}[ht]
\centering
\caption{Ranges of Parameters for Random Affine}
\SPACE
\begin{tabular}{c|rrc}
\hline
Type        & Lower Limit & Upper Limit & Distribution \\
\hline
Translate X & $-20\%$     & $20\%$     & NORMAL        \\
Translate Y & $-20\%$     & $20\%$     & NORMAL        \\
Scale       & $80\%$      & $160\%$    & NORMAL        \\
Rotate      & $-30\deg$   & $30\deg$   & NORMAL        \\
Shear       & $-5\deg$    & $5\deg$    & NORMAL        \\
\hline
\end{tabular}
\label{table:affine-params}
\end{table}

Random affine augmentation was employed in the training process. The affine transformation is a combination of five different transformations. Each transformation is controlled by a separate set of parameters. The list of applicable transformations and their probability distributions are listed in \Table{affine-params}, all parameters follow normal distribution within the given range.
This data augmentation technique is used to compensate for the limited amount of samples we are able to collect. It also helps to prevent the model from over fitting to scale and/or distortion.
If the transformed image cannot fill up the entire canvas, we have two different methods to fill it up. The simpler approach is to use a uniform average color for filling, another is mirroring the original image so the content looks consistent and meaningful. In practice, the mirroring option would consume too much resources (8 times more memory consumption) and would consequently limit the batch size to 1. Therefore, we adopted uniform solid color fill throughout all stages of training. In addition, the mask of valid pixels was preserved for each affine transform and re-applied for the loss function.

\begin{figure}[ht]
\includegraphics[width=\textwidth]{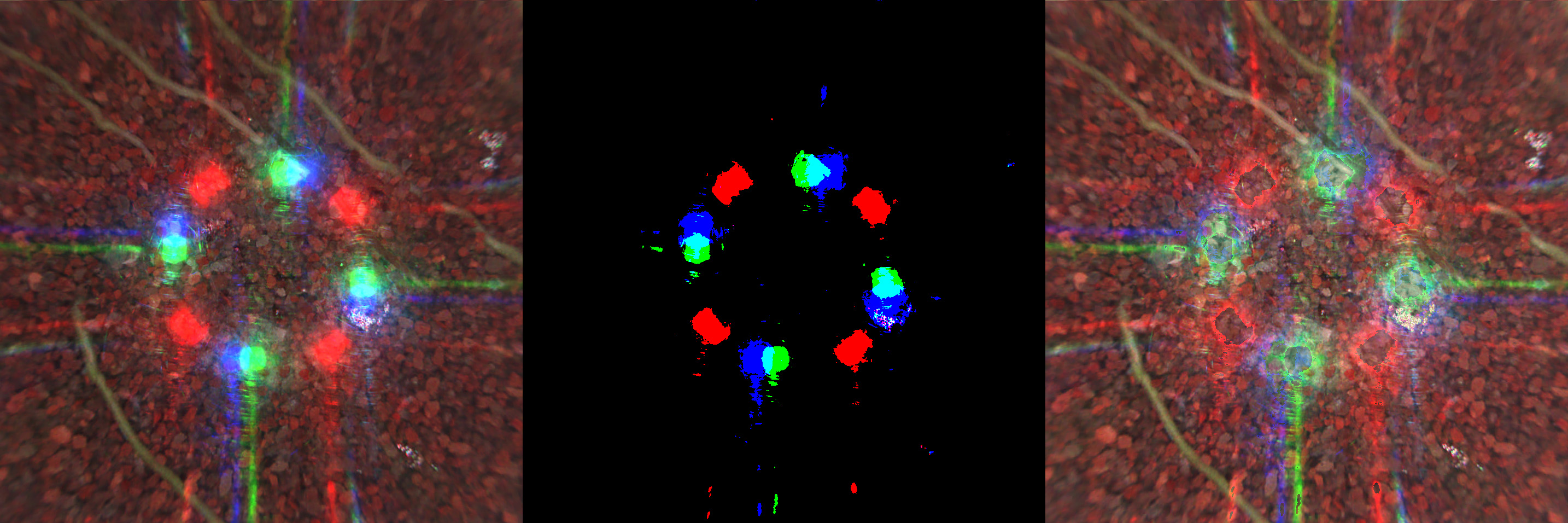}
\SPACE
\caption{Sample of otsu method preprocessing. Left: original image with LED light spots. Middle: Binary masks computed with otsu method. Right: enhanced image. All images have 8 channels, all displayed in pseudo RGB}
\label{fig:otsu}
\end{figure}

Another challenge is the bright LED spots found in our camera's results. These light spots (reflections of the LED lights) are supposed to be ignored and filled up by the model. In order to help our U-Net model learn to do this, we implemented Otsu's thresholding as an optional add-on to the data augmentation pipeline. The Otsu method searches for the threshold that minimizes the intra-class variation for both sides of the threshold. This method applies well on our case, which has a cluster of very bright pixels (the light spots) and another cluster of dim pixels (the soil and root pixels). As is shown in \Fig{otsu}, this method creates an accurate mask separating light spot pixels from useful data. Based on the mask, our framework then performs an interpolation based on other bandwidths of the same pixel which are not affected by the light spot and uses the interpolated value in place of the over-exposed value.

\subsection{Training Process}\label{sec:training}



Our processed dataset has 95 pairs of working samples, each of them consists of a
reference matrix of shape $286 \times 286 \times 299$ pixels (the ground truth) and an input matrix of shape $1024 \times 1024 \times 8$ pixels. These samples (95 total) were divided into 85 samples as training set and 10 samples as test set.
The list of samples used as test set was randomly generated upon initialization
of a training. No manual selection was involved in the process of splitting training set and test set.
However, it is worth noticing that there might be a minor overlap between
nearby samples. The percentage of overlap is generally less than 10\%, and our
splitting algorithm does \textbf{not} take care of the overlap.

We used a combination of several different loss function to train the model. The effective loss function is the weighted sum of the following functions:
\begin{itemize}
\item Mean average error (MAE), also known as $L_1$ distance.

\begin{equation*}
L_1\left[~\vec{x},~\vec{y}~\right] =
\sum_{n=1}^{N}\left|\vec{x}_n - \vec{y}_n\right|
\end{equation*}

\item Mean square error (MSE), also known as $L_2$ distance.

\begin{equation*}
L_2\left[~\vec{x},~\vec{y}~\right] =
\sum_{n=1}^{N}\left(\vec{x}_n - \vec{y}_n\right)^2
\end{equation*}

\item Delta-pixel error

This is a custom loss function that measures how different a pixel is against its neighboring pixels (in both width and height direction). The total error is the mean of all pixels' max absolute difference against their neighboring pixels. This loss function smooths out the noise introduced by the model.

\item Delta-bands error

This is a custom loss function that measures how different a pixel of a band is against its neighboring bands. The total error is the mean of all bands' max absolute difference against their neighboring bands. This loss function rewards smoothness of the band plot for each pixel.
\end{itemize}

It is worth mentioning that, weights for each component above were manually configured in each stage of training.
In pre-train phase, all above loss functions are used to derive the total loss. The delta-pixel and delta-bands error each has a weight of 4, the MSE and the MAE each has a weight of 1. In main training phase, we only use smooth L1 loss.

\begin{table}[ht]
\centering
\caption{Test Set Prediction Performance (Statistical)}
\SPACE
\begin{tabular}{c|c|c|cc}
\hline
Sampling Date & Box ID & Region & MAE Loss & MSE Loss \\
\hline
03/23/2023    & 77     & F1     & 0.036009 & 0.002377 \\
03/23/2023    & 77     & C5     & 0.029735 & 0.001592 \\
03/23/2023    & 69     & F5     & 0.032207 & 0.001837 \\
03/23/2023    & 77     & F5     & 0.036013 & 0.002481 \\
03/23/2023    & 77     & F7     & 0.032229 & 0.001981 \\
03/23/2023    & 69     & C1     & 0.029626 & 0.001686 \\
\hline
\end{tabular}
\label{table:test-set-results}
\end{table}

\begin{figure*}
\centering
\includegraphics[width=\textwidth]{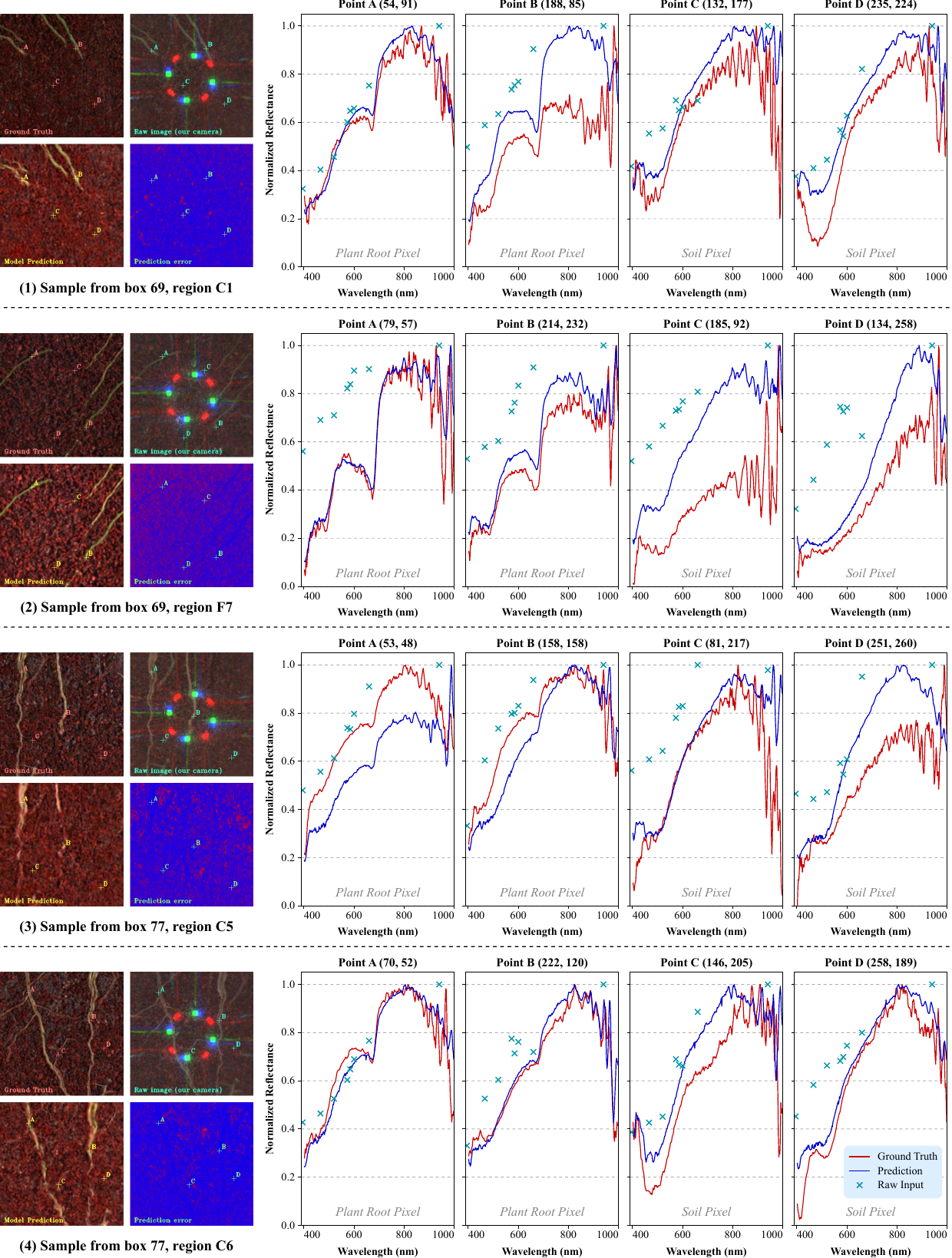}
\caption{
Samples of model prediction results, each comes with 4 manually selected sample points (A, B on plant root pixels; C, D on soil pixels).
Samples are selected with minimum bias.
Definition of "Prediction Error" is described in Sec.\ref{sec:results}
}
\label{fig:model-results}
\end{figure*}

The model was trained with a batch size of 5 for more than 1000 epochs each stage.

In the first stage, the model was fed with ground truth as input (pre-train). The ground truth (299 bands) data-points were first projected into the 8 bands corresponding to our camera's results, and then reshaped to mimic the dimensions of actual input data. This stage is designed to teach the network to learn to extend low dimension data into high dimensions without any interference. Random affine augmentation was enabled in this stage.

The model parameters from the first stage of training were then loaded back for the second (main) stage of training. In this stage, 3 out of 5 samples (each batch) were projected from the ground truth, just like what we did in the first stage. And the rest 2 samples are loaded from raw samples from our camera. Affine augmentation was also enabled in this process. This stage was added in order to make the network aware of LED light spots. The spots can appear in any shape, size and form factor with the help of affine transformation, they could also be completely absent since $60\%$ of the inputs used in this stage were converted directly from ground truth. This stage helps the network to gradually learn to deal with LED spots in our input dataset.

The mixture of ground truth inputs and raw inputs also ensured that the model "respects" the details of its input. In our previous trials, the model over-processed the input and removed the details entirely due to pixel shifts caused by alignment errors. This small misalignment confused the network and issued a punishment when the model generated sharp and detailed predictions. This misalignment can be found by comparing point B in sample (3) of \Fig{model-results}. In ground truth, point B sits on the left side of the root, while in the raw input, point B sits on the right side. This mismatch can also be spotted from the prediction error, the blue curve and red curve concentrate on different sides of the sample point.



\subsection{Training Result and Model Performance}

Due to the high volume of data expected in both spatial and spectral dimensions, it is not straightforward to demonstrate the overall performance in generalized metrics. Instead, we randomly picked a set of test samples and the spectrum of select pixels in these samples to demonstrate the training result (shown in \Fig{model-results}).

\label{sec:results}
In the following analysis and "Prediction Error" graphs shown in \Fig{model-results}, we used the spectral angular difference to quantize prediction. It is defined as:

$$
\begin{aligned}
\text{Let} ~~ \vec{p}_1 = \text{IMG}_\text{ground truth}  (x, y) &, \quad\text{and} ~~ \vec{p}_2 = \text{IMG}_\text{prediction} (x, y)\\[2mm]
E(x, y) = \arccos&{\left( \dfrac{\vec{p}_1 \cdot \vec{p}_2}{\sqrt{|\vec{p}_1|^2 \cdot |\vec{p}_2|^2}} \right)}
\end{aligned}
$$

This per-pixel error quantization treats pixels as high dimensional vectors (299 dimensions in our case), and calculates per-pixel angular difference for each spatial location $(x, y)$ between ground truth $\vec{p}_1$ and model prediction $\vec{p}_2$. In this way, the error compares the \textbf{shape} of spectral dimension and ignores difference in pixel intensity.

As is shown in the figure, the reconstruction model was able to bring out the details of the spectrum of a root pixel, especially the curved region near $700$nm. The model also did well in reconstructing average soil pixels, which are expected to look like a nice Gaussian distribution because it is composed by a mixture of many different ingredients.

With the help of the data processing and training techniques shown in in Sec.\ref{sec:training}, our model performed well in handling light spots. Although the output is relatively blurry in spatial dimensions, a significant amount of plant root features were conserved in the spectral dimension and showed a high relevance towards the ground truth.

\label{sec:quant-analysis}

\begin{figure}
\centering
\includegraphics[width=\textwidth]{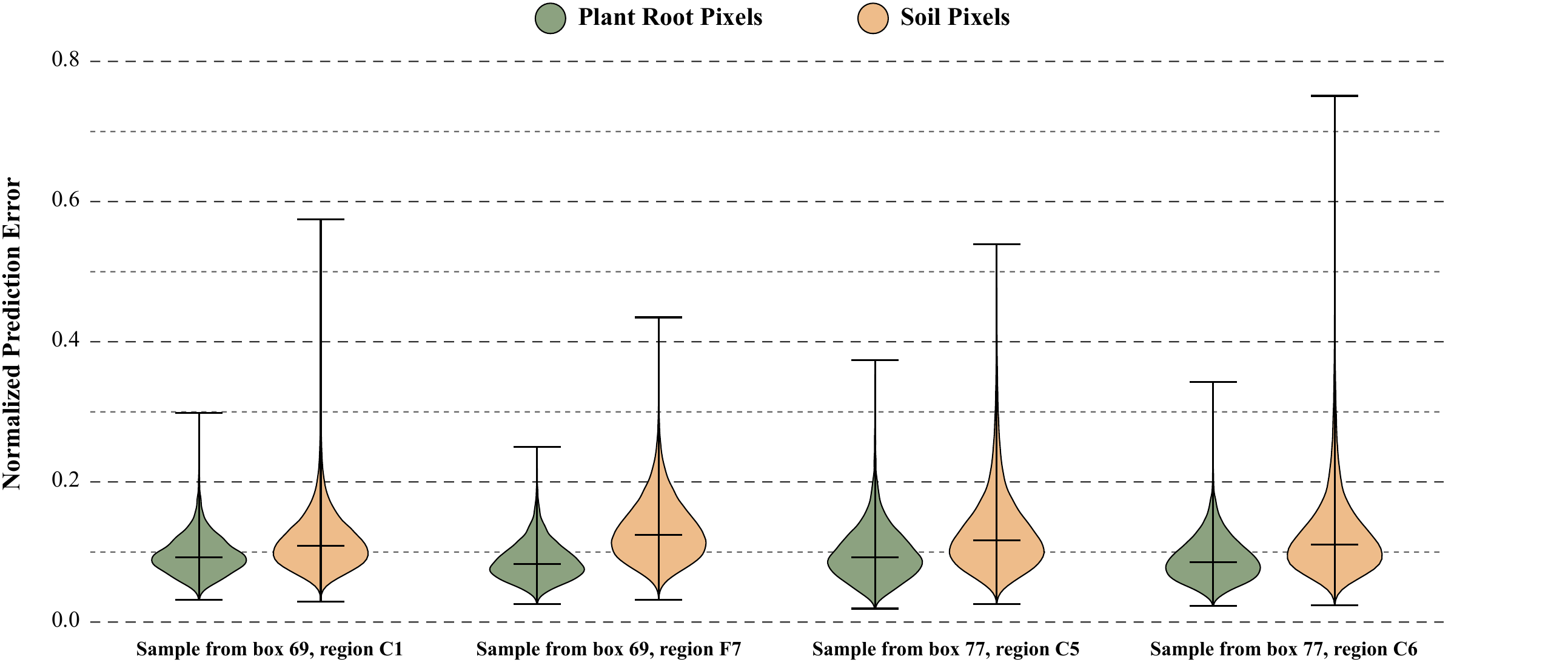}
\caption{
Statistical model prediction error. Error bars for root and soil pixels are shown separately for each individual sample.
Angular error was divided by a factor of $\frac{\pi}{2}$, normalizing the max possible error to $1.0$.
}
\label{fig:stats}
\end{figure}

In addition to the above qualitative analysis, thanks to the segmentation model\cite{chang2023hyperpri} provided by Spencer et al., the prediction error was categorized into plant root pixels and soil pixels respectively. As is shown in \Fig{stats}. derived prediction error from 4 prediction images are each divided into root pixels and soil pixels. The figure indicates that our model performed slightly better when handling plant root pixels.

\subsubsection{Limitations}
As shown in Sec.\ref{sec:quant-analysis}, our model did not perform as great then handling soil pixels. This is partially caused by the difficulty to accurately match up high spatial frequency details of fine-grained soil pixels (see \Fig{aligned-image}.D). In addition to the error introduced by template matching algorithms, we also noticed that some soil grains did not stay in the same location before and after we retrieved the rhizobox from the cabinet for the reference HSI camera. Although done carefully, the transportation inevitably caused some soil grains to move around, causing a mismatch between ground truth and our camera's data. This might also be a significant contributing factor for the blurry output of our model.

Although the model showed certain capabilities of providing extended spectral information based on its 8-band input, it is worth mentioning that we do not expect the same model to work as well for unknown plants or unknown variants. The reason we are able to use only 8 bands to reconstruct the entire spectrum is the network learns the types and states of a pixel in 8 different bands and internally maps it back to the same known state of a high resolution spectrum distribution. This process is purely based on the previous "knowledge" stored inside the model. It should not be able to be transferred to unknown plant species nor different soil types.

\section{Conclusion}

In this work, we demostrated the feasibility of a cost-efficient approach to build a compact sized hyper-spectral camera. We performed experiments on spectrum reconstruction from a reduced spectral image back into a full-sized hyper-spectral image. The reconstruction results showed higher relevance for plant root pixels, but did not perform as well on soil pixels. Such a result indicates that additional, useful data can indeed be obtained by our active illumination camera setup.

For the next step, we will capture data from real minirhizotrons and validate our model's ability to transfer its knowledge to those data. This is expected to be more challenging because the plants might be of a different species, and the type of soil might be drastically different than the rhizoboxes from which we obtained data for training.

\begin{backmatter}
\bmsection{Funding}
Placeholder: will be replaced in submission build.


\bmsection{Disclosures}
The authors declare no conflicts of interest.










\bmsection{Data availability} Data underlying the results presented in this paper are not publicly available at this time but may be obtained from the authors upon reasonable request.







\end{backmatter}

\bibliography{paper}

\end{document}